\pdfoutput=1
\documentclass[10pt,twocolumn,letterpaper]{article}

\usepackage[pagenumbers]{cvpr}   % arXiv version: non-anonymous, with page numbers

\usepackage{booktabs}
\usepackage{multirow}

\usepackage{xspace}
\newcommand{\method}{FastD4RT\xspace}

\newtheorem{theorem}{Theorem}
\newtheorem{assumption}{Assumption}

\definecolor{cvprblue}{rgb}{0.21,0.49,0.74}
\definecolor{oursPurple}{RGB}{232,222,242}
\usepackage[breaklinks,colorlinks,allcolors=cvprblue]{hyperref}
\usepackage{algorithm}
\usepackage{algpseudocode}
\usepackage[table]{xcolor}
\usepackage{multirow}
\usepackage{pifont}

\def\confName{CVPR}
\def\confYear{2027}

\title{\method: Training-Free Acceleration of Dynamic 4D \\ Reconstruction and Tracking}

\author{
Meng Bi$^{1}$ \quad Yili Jin$^{2}$ \quad Kaiyuan Hu$^{2,3}$ \quad Xize Duan$^{3}$ \quad Jiaxin Huang$^{1}$ \quad Hong Huang$^{4}$ \quad Xue Liu$^{1,3}$ \\[0.6em]
{\normalsize $^{1}$MBZUAI \quad $^{2}$MirrorSpace \quad $^{3}$McGill University \quad $^{4}$City University of Hong Kong}
}

\begin{document}
\twocolumn[{%
\renewcommand\twocolumn[1][]{#1}%
\maketitle
\begin{center}
    \centering
    \captionsetup{type=figure}
    \includegraphics[width=0.95\textwidth]{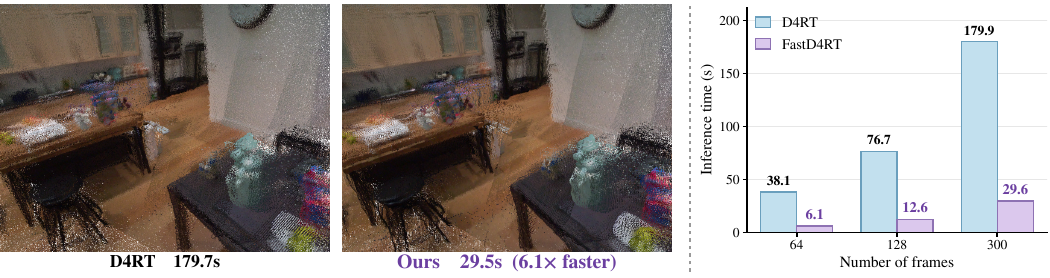}
    \captionof{figure}{\textbf{\method} accelerates D4RT~\cite{zhang2025d4rt} without retraining. \textbf{Left:} On a 300-frame video, \method runs much faster than D4RT with nearly identical accuracy. \textbf{Right:} Averaged over videos of each length, \method greatly reduces the inference time of D4RT.}
    \label{fig:teaser}
\end{center}%
}]
\begin{abstract}
Query-based 4D reconstruction models such as D4RT encode a video once into a global scene representation and obtain outputs by decoding independent point queries. However, inference is often influenced by dense outputs and long videos. Although D4RT separates encoding from decoding, a dense output still requires one decoder query for every pixel of every frame, and a long video is still encoded as many overlapping segments. In this work, we analyze the inference cost of D4RT and show that a large part of it is redundant. In particular, we find that the later decoder blocks attend to a small set of tokens shared by all queries. Based on these findings, we propose FastD4RT, a training-free framework that accelerates D4RT. By exploiting the smoothness of the outputs over time and across pixels, FastD4RT encodes and decodes only at sparse anchors and recovers the remaining outputs via interpolation. It decodes the remaining queries against a shared token set selected once per segment. We further characterize the errors introduced by both components. Experimental results show that FastD4RT accelerates D4RT by about 5 times on average across all its tasks with competitive accuracy, and further improves accuracy on long videos. Our code is available at \url{https://github.com/Lycheemeng/fastd4rt}.

\end{abstract}
\section{Introduction}
\label{sec:intro}

Reconstructing the geometry and motion of a dynamic scene from a monocular video is critical for applications such as robotics~\cite{bharadhwaj2024track2act,wen2024atm} and augmented reality~\cite{luo2020consistent}. Recent feed-forward models~\cite{wang2024dust3r,zhang2025monst3r,wang2025cut3r,wang2025vggt,yang2025fast3r,wang2025pi3} replace per-scene optimization~\cite{kopf2021robust,zhang2022casualsam,li2025megasam} with a single network pass and have made this task far more practical. Typically, these models predict a dense output, such as a point map or a depth map, for every input frame through task-specific heads. However, such per-frame outputs do not directly describe how a point moves over time, so 4D tasks such as 3D point tracking still rely on separate modules~\cite{xiao2024spatialtracker,ngo2025delta,xiao2025spatialtrackerv2}, and every frame has to be fully decoded even when only a few points are needed. To address this, query-based 4D reconstruction models such as D4RT~\cite{zhang2025d4rt} decouple encoding from decoding. A transformer encoder maps the whole video to a global scene representation once, and a lightweight decoder then answers independent queries: given a 2D point in a source frame, it returns the 3D position of that point at any target time and in any camera. In this way, different 4D outputs are obtained from a single model by choosing different queries, and only the queried outputs need to be decoded. Since the video is encoded only once and every query is decoded independently, D4RT is designed to be efficient and is reported to be much faster than prior 3D trackers.

\begin{figure}[t]
    \centering
    \includegraphics[width=\linewidth]{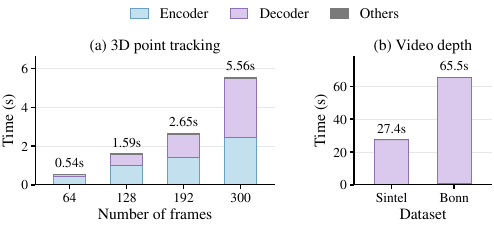}
    \caption{\textbf{Component-wise analysis of D4RT inference time.} Numbers above the bars give the total runtime per video. (a) In 3D point tracking, the encoder dominates on short videos, while the decoder takes a growing share as video gets longer. (b) In video depth estimation, the decoder accounts for almost all the runtime, suggesting that dense decoding is highly inefficient.}
    \label{fig:runtime}
\end{figure}

However, this design has a key limitation: D4RT still encodes every frame and decodes every output densely, even when most of this computation is redundant. A full-length 3D track needs one query per frame, and a dense depth map needs one query per pixel. Meanwhile, the encoder only processes clips of a fixed length, so a long video has to be encoded as many overlapping segments.  As shown in Fig.~\ref{fig:runtime}, the encoder and the decoder each take a large share of the runtime in 3D tracking, while the decoder alone accounts for almost all of the runtime in dense depth estimation. This creates a severe bottleneck and limits the scalability of D4RT to long videos and dense outputs.

To understand where this redundancy comes from, we present a detailed analysis of the inference of D4RT. Our analysis reveals that the redundancy lies at three levels. First, most decoder queries are redundant. A 3D trajectory usually changes smoothly over time, and a depth map is decoded from a scene representation with a much coarser grid of tokens, so most outputs can be recovered from their neighbors. In fact, D4RT already uses this property for camera parameters, which are estimated from a coarse grid of queries, but not for the other outputs. Second, the encoded segments are redundant. Adjacent frames of a video are highly similar, yet a long video is encoded densely with overlapping segments. More importantly, the attention inside each query is also redundant. We find that in the later decoder blocks, all queries attend to almost the same small set of tokens, although each query still attends to the whole scene representation. This suggests that the later blocks read a shared summary of the scene rather than searching for information specific to each query. 

Motivated by these observations, we propose \method, a training-free framework that accelerates D4RT. To reduce the number of computations, we propose anchor-based sparse inference, which decodes each output only at sparse anchors and interpolates the rest. We further derive an upper bound for the interpolation error. To reduce the cost of each query, we propose shared-token decoding, letting all queries in the later decoder blocks attend only to a small shared set of tokens.

Furthermore, the same idea enables faster and more accurate inference on long videos. For long videos, \method encodes only sparse anchor frames, so the video is split into fewer segments. Since every segment adds one more alignment to the chain, fewer segments also mean less accumulated error, making \method faster and more accurate on long videos.

We evaluate \method on the tasks and benchmarks of D4RT to demonstrate its effectiveness on both sparse and dense outputs. Since the official model is not publicly available, all experiments are built on the community OpenD4RT weights. Our comprehensive experiments show that \method consistently accelerates D4RT while preserving its accuracy. For instance, \method is 6.1 times faster on dense prediction with comparable accuracy, and produces about 5 times more full-video 3D tracks at every target frame rate.

Our contributions are summarized as follows:
\begin{itemize}
    \item We provide a detailed analysis of the inference cost of D4RT, and identify where the computation is redundant. In particular, we find that most outputs change smoothly and can be recovered from sparse anchors, that the later decoder blocks attend to a small set of tokens shared by all queries, and that more segments lead to more accumulated error on long videos.
    \item We introduce \method, a training-free framework that combines anchor-based sparse inference with shared-token decoding, whose speed-ups compose multiplicatively. We further derive upper bounds for the errors introduced by both components.
    \item We evaluate \method, across the tasks and benchmarks of D4RT. Results indicate that \method accelerates D4RT by 2.1--6.2 times with comparable accuracy, produces about 5 times more 3D tracks at every target frame rate, and further improves accuracy on long videos.
\end{itemize}
\section{Preliminary and Challenges}
\label{sec:preliminary}

\subsection{Query-Based 4D Reconstruction}
\label{sec:prelim}

Query-based 4D reconstruction recovers the geometry and motion of a dynamic scene from a video $\mathcal{V}$ with an encoder and a decoder. Unlike feed-forward models that predict a dense output for every input frame~\cite{wang2025cut3r,wang2025vggt}, it first encodes the video into a scene representation, and then obtains each output by querying this representation. In D4RT~\cite{zhang2025d4rt}, the encoder $\mathcal{E}$ is a ViT-g with interleaved frame-wise and global attention, which maps a clip to a global scene representation
\begin{equation}
    \mathbf{F} = \mathcal{E}(\mathcal{V}) \in \mathbb{R}^{N\times d},
    \label{eq:encoder}
\end{equation}
where $N$ is the number of tokens.\\
A query is a 2D point together with three time indices,
\begin{equation}
    \mathbf{q} = (u, v, t_{\text{src}}, t_{\text{tgt}}, t_{\text{cam}}),
    \label{eq:query}
\end{equation}
where $(u,v)\in[0,1]^2$ is a location in frame $t_{\text{src}}$. Given $\mathbf{q}$, the decoder $\mathcal{D}$ returns the 3D position of this point at time $t_{\text{tgt}}$, expressed in the camera of frame $t_{\text{cam}}$:
\begin{equation}
    \mathbf{P}=\mathcal{D}(\mathbf{q},F)\in\mathbb{R}^{3}.
    \label{eq:decoder}
\end{equation}
The decoder also predicts the visibility and the 2D position of the point, which are used for tracking. It consists of eight blocks of cross-attention, so every query can be decoded independently.

The core of D4RT lies in this query interface: different tasks differ only in the queries they issue. The 3D track is obtained by querying the same point at every target time, and a depth map is obtained by querying every pixel of a frame. Camera parameters are solved from the 3D points of a coarse grid of queries in each frame. As a result, tasks that previously required separate models or dedicated output heads are handled by a single encoder–decoder, which distinguishes D4RT from prior 4D reconstruction methods.

\noindent\textbf{Long-video inference.} D4RT is trained on short clips. For longer videos, it splits the video into $K$ overlapping segments and encodes each segment once. Each segment is aligned to its predecessor by Sim(3) transform estimated on their overlap~\cite{umeyama1991least}, and the transforms are chained to the first segment. This follows the first stage of VGGT-Long~\cite{deng2026vggt}, without its loop detection and global optimization. D4RT does not specify how tracks are carried across segments; in our implementation, a query starts in the segment closest to its source frame and is handed over to the next segment through its decoded 2D position in the overlap.

\subsection{Challenges in D4RT Inference}
\label{sec:prelim_challenge}
Since D4RT encodes each segment once and decodes every query independently, its inference cost separates into two terms and can be approximated as,
\begin{equation}
    C \approx K\,C_{\mathcal{E}} + N_q\,C_{\mathcal{D}},
    \label{eq:cost}
\end{equation}
where $C_{\mathcal{E}}$ is the cost of encoding one segment, $N_q$ is the number of decoder queries, and $C_{\mathcal{D}}$ is the cost of decoding one query. The first term grows with the length of the video, since $K$ increases with $T$. The second term grows with the size of the output: tracking $M$ points over $T$ frames issues $N_q=MT$ queries, and a dense output at resolution $H\times W$ issues $N_q=HWT$ queries.

However, both terms in Eq.~\eqref{eq:cost} remain large in practice. As shown in Fig.~\ref{fig:runtime}, encoding the segments and decoding the track queries take a comparable share of the runtime in 3D tracking, while the decoder accounts for almost all of the runtime in dense depth estimation. This issue is exacerbated on long videos, where more segments are encoded and every track is queried at more frames. Moreover, each query attends to all $N$ tokens in every decoder block, so even a single query is costly. Consequently, the core challenge lies in reducing both the number of computations, $K$ and $N_q$, and the cost $C_{\mathcal{D}}$ of each query.
\begin{figure}[t]
    \centering
    \includegraphics[width=\linewidth]{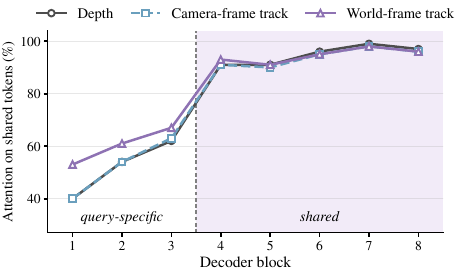}
    \caption{Fraction of the attention captured by a shared token set in each decoder block.}
    \label{fig:shared_ratio}
\end{figure}

\section{Methodology}
\label{sec:method}

In this section, we propose FastD4RT, motivated by a key limitation of D4RT inference: it encodes every frame and decodes every query densely. We show that most outputs can be recovered from a few sparse anchors, and that the later decoder blocks only need a small set of tokens shared by all queries. Building on these findings, we present anchor-based sparse inference to reduce the number of encoded segments and decoded queries, and shared-token decoding to reduce the cost of each query, together with their complexity. Finally, we describe the overall framework and derive its error bounds.

\subsection{Problem Reformulation}
\label{sec:method_problem}

D4RT aims to make 4D reconstruction efficient by encoding a video once and answering independent queries. However, we find that its inference computes far more than the outputs require. Our central claim is that this extra computation comes from redundancy: most outputs change smoothly and carry little new information beyond their neighbors, and the later decoder blocks attend to the same small set of tokens for all queries. We reduce $K$ and $N_q$ in Eq.~\eqref{eq:cost} with anchor-based sparse inference in Sec.~\ref{sec:method_sparse}, and reduce $C_{\mathcal{D}}$ with shared-token decoding in Sec.~\ref{sec:method_shared}.

\begin{figure*}[t]
    \centering
    \includegraphics[width=\linewidth]{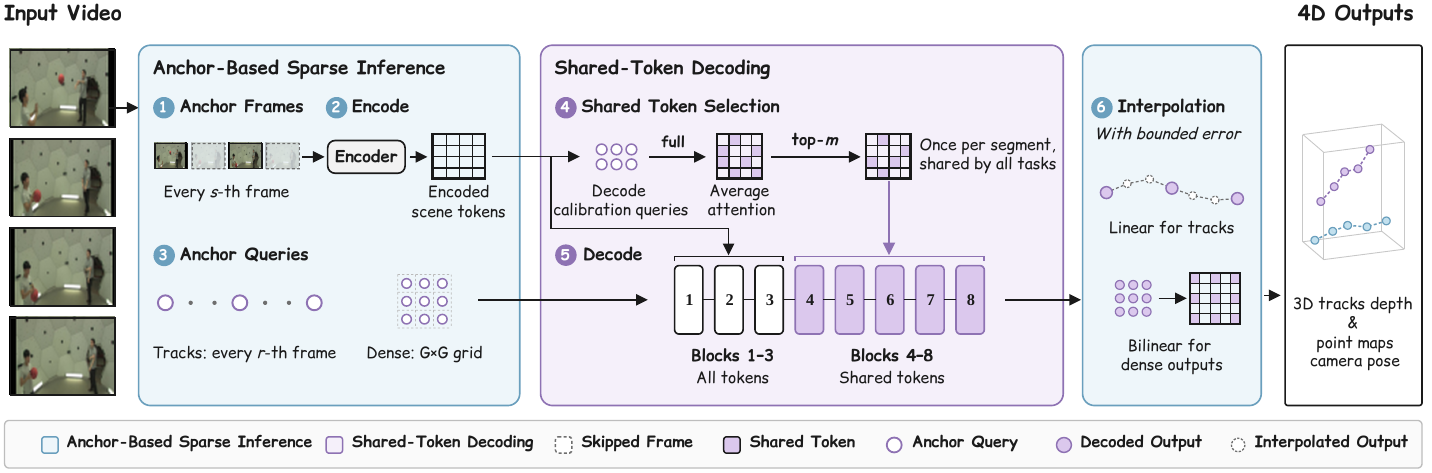}
    \caption{\textbf{Overview of \method.} Given a video, \method encodes only sparse anchor frames, so that a long video is split into fewer segments. For each segment, a small set of calibration queries selects a shared token set once. Each task then issues queries only at sparse anchors, which attend to all tokens in the early decoder blocks and only to the shared tokens in the later blocks. Finally, the remaining outputs are recovered by interpolation, and adjacent segments are aligned.}
    \label{fig:framework}
\end{figure*}

\subsection{Anchor-Based Sparse Inference}
\label{sec:method_sparse}

As discussed in Sec.~\ref{sec:prelim_challenge}, D4RT decodes every query and encodes every frame. But most of its outputs change smoothly and can be recovered from their neighbors. Therefore, we decode each output only at anchors placed every $\rho_j$ steps along each axis $j$ where it is smooth, together with the last point, and recover the rest by linear interpolation. For a point $t$ between two neighboring anchors $a<t<b$,
\begin{equation}
    \hat{\mathbf{y}}(t) = \frac{b-t}{b-a}\,\mathbf{y}(a) + \frac{t-a}{b-a}\,\mathbf{y}(b).
    \label{eq:interp}
\end{equation}

For 3D tracks and camera poses, we place anchors along time with spacing $\rho_t=r$ and interpolate the rotations along the geodesic. They are decoded at only one in every $r$ frame, which reduces their queries by a factor of $r$. For depth maps and point clouds, the decoder accepts continuous coordinates $(u,v)$, so we decode them on a coarse $G\times G$ grid with spacing $\rho_u=H/G$ and $\rho_v=W/G$, and upsample the result to $H\times W$ with bilinear interpolation. This reduces the queries per frame from $HW$ to $G^2$.

The same idea applies to the encoder. For long videos, the encoder takes only every $s$-th frame, so the number of segments drops by about a factor of $s$. Since every segment adds one more alignment to the chain, fewer segments also reduce the error accumulated over a long video.

\noindent\textbf{Complexity.} Anchor-based sparse inference reduces the number of decoded queries in Eq.~\eqref{eq:cost} from $N_q$ to $N_q/\gamma$, with $\gamma=r$ for tracks and camera poses and $\gamma=HW/G^2$ for dense outputs. For long videos, it reduces the number of encoded segments from $K$ to about $K/s$. With $r=3$ and $s=2$ in our experiments, tracks require 3 times fewer queries, and long videos about 2 times fewer segments.

\subsection{Shared-Token Decoding}
\label{sec:method_shared}

We introduce shared-token decoding, which reduces the cost of each remaining query while keeping the early decoder blocks unchanged.

\noindent\textbf{Attention analysis.} We first examine how the decoder uses the scene representation. For each decoder block, we select a small set of tokens with the largest mean attention over calibration queries, and measure how much of the attention of held-out queries falls on them. As shown in Fig.~\ref{fig:shared_ratio}, in the first three blocks, this set captures only part of the attention, so each query attends to its own tokens. From the fourth block on, it captures almost all of the attention for every type of query. This suggests that the early blocks locate the information for each query, while the later blocks read a shared summary of the scene. Motivated by this observation, we restrict the later blocks to a shared set of tokens.

\noindent\textbf{Shared token selection.} For each segment, we first decode a small set of calibration queries $\mathcal{C}$ with full attention. Let $\mathbf{a}(\mathbf{q})\in\mathbb{R}^N$ denote the attention of query $\mathbf{q}$ over the $N$ tokens in a head. For each head of block after $l_0$, we average the attention over the calibration queries and keep the $m$ tokens with the largest mean attention:
\begin{equation}
    \mathcal{T} = \operatorname{top-}\!m\Big(\frac{1}{|\mathcal{C}|}\sum_{\mathbf{q}\in\mathcal{C}}\mathbf{a}(\mathbf{q})\Big).
    \label{eq:shared}
\end{equation}

\noindent\textbf{Shared-token attention.} The queries are unchanged, while the keys and values of the blocks after $l_0$ are restricted to the shared tokens:
\begin{equation}
    \mathbf{z}_{l+1} =
    \begin{cases}
        \mathcal{D}_l(\mathbf{z}_l, \mathbf{F}), & l \le l_0, \\
        \mathcal{D}_l(\mathbf{z}_l, \mathbf{F}_{\mathcal{T}}), & l > l_0,
    \end{cases}
    \label{eq:shared_attn}
\end{equation}
where $\mathcal{D}_l$ is the $l$-th decoder block and $\mathbf{z}_l$ is the query feature at block $l$. Crucially, the token set is selected once per segment and shared by all queries. Since the calibration queries cover all tasks, the same token set serves every task.

\noindent\textbf{Complexity.} Let the decoder have width $d$. Each block costs $\mathcal{O}(Nd)$ for cross-attention and $\mathcal{O}(d^2)$ for projections and MLP, so $C_{\mathcal{D}}=\mathcal{O}\big(L(Nd+d^2)\big)$ in Eq.~\eqref{eq:cost}. Shared-token decoding reduces this to $\mathcal{O}\big(l_0Nd+(L-l_0)md+Ld^2\big)$, at the overhead of decoding $|\mathcal{C}|$ calibration queries per segment. With $m=512$ in our experiments, cross-attention in the last $L-l_0$ blocks is about 24 times cheaper.

\subsection{Overall Framework}
\label{sec:method_framework}

Based on the above analysis of anchor-based sparse inference and shared-token decoding, we propose FastD4RT, a training-free framework for accelerating query-based 4D reconstruction, as illustrated in Fig.~\ref{fig:framework}. FastD4RT follows the same segment-based pipeline as D4RT. On the encoder side, a long video is subsampled with stride $s$, so that fewer segments are encoded. On the decoder side, calibration queries select a shared token set once per segment, and the queries of each task are decoded only at sparse anchors, with the last blocks attending only to the shared tokens. The remaining outputs are then recovered by interpolation and chained across segments. FastD4RT applies this process to every segment, and detailed procedures are provided in Algorithm~\ref{alg:fastd4rt} and Appendix~\ref{app:framework}. Since anchors reduce the number of decoded queries and shared tokens reduce the cost of each one, the two components compose multiplicatively.

\subsection{Theoretical Error Bounds}
\label{sec:method_bound}

FastD4RT introduces two sources of error relative to D4RT: (i) the interpolation error of anchor-based sparse inference, and (ii) the shared-token error of shared-token decoding. We bound each of them below.

\noindent\textbf{Interpolation error.} Let $E_{\text{int}}$ be the error of interpolating at the anchors. For dense outputs, let $\Delta$ be the relative depth difference between neighboring grid points, and $\ell$ the difference of the output between them, which is the depth difference for depth maps and the 3D distance for point clouds.

\begin{assumption}[Smooth outputs]
\label{assu:smooth}
Between neighboring anchors, every point tracked by D4RT moves with an acceleration of at most $\alpha$, and every surface predicted by D4RT is planar.
\end{assumption}

\begin{theorem}[Interpolation error]
\label{thm:interp}
Under Assumption~\ref{assu:smooth}, $E_{\text{int}}\le\frac{r^2}{8}\alpha$ for 3D tracks, and $E_{\text{int}}\le\frac{1}{2}\Delta\,\ell$ for depth maps and point clouds.
\end{theorem}

The interpolation error grows with the anchor spacing, and vanishes for static points and fronto-parallel surfaces.

\noindent\textbf{Shared-token error.} Shared-token decoding introduces error only through the attention heads in the blocks after $l_0$. So we measure this error at the output of each head.    Let $\mathbf{o}$ and $\tilde{\mathbf{o}}$ be the outputs of a head with full attention and with attention restricted to the shared tokens, whose difference $E_{\text{sh}}=\|\tilde{\mathbf{o}}-\mathbf{o}\|$ is the shared-token error. Let $\epsilon$ be attention mass outside the shared tokens, and $V$ be the largest norm of the values over all tokens.

\begin{theorem}[Shared-token error]
\label{thm:shared}
For any query and any attention head in the blocks after $l_0$, the shared-token error satisfies $E_{\text{sh}}\le 2\epsilon V$.
\end{theorem}

The shared-token error is mainly governed by the attention mass $\epsilon$ outside the shared tokens, which Fig.~\ref{fig:shared_ratio} shows to be small from the fourth block on.

\noindent\textbf{Takeaways.} Theorems~\ref{thm:interp} and~\ref{thm:shared} show that the errors of FastD4RT satisfy
\begin{equation}
\begin{aligned}
    E_{\text{int}} &= O(r^2) && \text{for tracks},\\
    E_{\text{int}} &= O\big((H/G)^2\big) && \text{for dense outputs},\\
    E_{\text{sh}} &= O(\epsilon).
\end{aligned}
\label{eq:takeaways}
\end{equation}
This result shows that both accelerations are nearly lossless where they are applied. The interpolation error depends only on the output smoothness, so most queries can be skipped for smooth motion and surfaces, and the shared-token error depends only on $\epsilon$, which is small in the blocks after $l_0$. Since interpolation does not amplify anchor errors, the two errors do not compound, and FastD4RT stays close to D4RT whenever its outputs are smooth between anchors.

\begin{table*}[t]
    \caption{\textbf{4D reconstruction and tracking} -- 3D tracking on TAPVid-3D with tracks predicted in local camera coordinates (left) and world coordinates (right), without (top) and with (bottom) ground-truth intrinsics. AJ, APD (= APD$_{3D}$) and OA are in \%. Speed-up is relative to D4RT. \method achieves accuracy comparable to or better than D4RT while being substantially faster.}
    \centering
    \footnotesize
    \setlength{\tabcolsep}{5pt}
    \begin{tabular}{c l ccc ccc ccc cc cc}
        \toprule
        \multirow{3}{*}[-1ex]{\textbf{\begin{tabular}[c]{@{}c@{}}w/ GT\\intrin.\end{tabular}}}
        & \multirow{3}{*}[-1ex]{\textbf{Method}}
        & \multicolumn{9}{c}{\textbf{Camera frame}} & \multicolumn{4}{c}{\textbf{World frame}} \\
        \cmidrule(lr){3-11} \cmidrule(lr){12-15}
        & & \multicolumn{3}{c}{\textbf{DriveTrack}} & \multicolumn{3}{c}{\textbf{ADT}} & \multicolumn{3}{c}{\textbf{PStudio}}
          & \multicolumn{2}{c}{\textbf{DriveTrack}} & \multicolumn{2}{c}{\textbf{ADT}} \\
        \cmidrule(lr){3-5} \cmidrule(lr){6-8} \cmidrule(lr){9-11} \cmidrule(lr){12-13} \cmidrule(lr){14-15}
        &
            & AJ$\uparrow$ & APD$\uparrow$ & OA$\uparrow$
            & AJ$\uparrow$ & APD$\uparrow$ & OA$\uparrow$
            & AJ$\uparrow$ & APD$\uparrow$ & OA$\uparrow$
            & APD$\uparrow$ & L1$\downarrow$ & APD$\uparrow$ & L1$\downarrow$ \\
        \midrule
        \multirow{3}{*}{\ding{55}}
        & D4RT     & 1.24 & 2.07 & 82.85 & 2.26 & 4.65 & 77.30 & 6.47 & 10.99 & 82.63
            & 3.10 & 8.34 & 34.79 & 1.78 \\
        & OpenD4RT & 1.23 & 2.04 & 82.88 & 2.27 & 4.67 & 77.31 & 6.48 & 10.99 & 82.70
            & 3.07 & 8.34 & 34.89 & 1.78 \\
        & \textbf{\method (Ours)} & 1.25 & 2.15 & 81.71 & 3.01 & 5.46 & 78.03 & 6.61 & 11.06 & 83.60
            & 3.16 & 8.48 & 39.30 & 1.56 \\
        \midrule
        \midrule
        \multirow{3}{*}{\ding{51}}
        & D4RT     & 6.04 & 10.14 & 82.85 & 4.15 & 8.21 & 77.30 & 11.29 & 18.87 & 82.63 & 6.81 & 6.38 & 31.67 & 1.96 \\
        & OpenD4RT & 5.98 & 10.04 & 82.88 & 4.09 & 8.10 & 77.31 & 11.15 & 18.68 & 82.70 & 6.80 & 6.38 & 31.50 & 1.96 \\
        & \textbf{\method (Ours)} & 6.13 & 10.12 & 81.71 & 4.43 & 7.89 & 78.03 & 12.10 & 19.78 & 83.60
            & 7.93 & 6.54 & 33.16 & 1.99 \\
        \midrule
        \specialrule{\lightrulewidth}{0pt}{0pt}
        \rowcolor{oursPurple}
        & \textbf{Speed-up}\rule[-0.9ex]{0pt}{3ex} & \multicolumn{3}{c}{\textbf{2.8$\times$}} & \multicolumn{3}{c}{\textbf{5.3$\times$}} & \multicolumn{3}{c}{\textbf{4.2$\times$}}
                   & \multicolumn{2}{c}{\textbf{3.1$\times$}} & \multicolumn{2}{c}{\textbf{5.8$\times$}} \\
        \specialrule{\heavyrulewidth}{0pt}{0pt}
    \end{tabular}
    \label{tab:tracking}
\end{table*}

\section{Experiments}
\label{sec:exp}

This section evaluates \method through extensive experiments on the tasks and benchmarks of D4RT. Following D4RT, we first evaluate 4D reconstruction and tracking in Sec.~\ref{sec:exp_4d}, and then 3D reconstruction in Sec.~\ref{sec:exp_3d}. We further examine how \method behaves as videos get longer in Sec.~\ref{sec:long_video}. All experiments are repeated three times, and we report the mean results.

%% ---------------------------------------------------------------- Tables

\begin{table}[t]
    \caption{\textbf{3D tracking throughput} -- We measure the maximum number of full-video 3D point tracks that different methods can produce while maintaining a given FPS target on a single H20 GPU. Speed-up is relative to D4RT.}
    \centering
    \small
    \begin{tabular}{l rrrr}
        \toprule
        & \multicolumn{4}{c}{Max.\ track count @ target FPS} \\
        \cmidrule(lr){2-5}
        Method & 60 FPS & 24 FPS & 10 FPS & 1 FPS \\
        \midrule
        D4RT & 1,342 & 4,173 & 10,360 & 107,761 \\
        OpenD4RT & 0 & 475 & 1,719 & 20,926 \\
        \textbf{\method (Ours)} & 6,653 & 19,109 & 48,641 & 519,872 \\
        \midrule
        \specialrule{\lightrulewidth}{0pt}{0pt}
        \rowcolor{oursPurple}
        \textbf{Speed-up}\rule[-0.9ex]{0pt}{3ex} & \textbf{5.0$\times$} & \textbf{4.6$\times$} & \textbf{4.7$\times$} & \textbf{4.8$\times$} \\
        \specialrule{\heavyrulewidth}{0pt}{0pt}
    \end{tabular}
    \label{tab:throughput}
\end{table}

\begin{table*}[t]
    \caption{\textbf{Video depth and point cloud estimation.} \method achieves accuracy comparable to D4RT under both scale-only (S) and scale-and-shift (SS) alignments while being substantially faster. Speed-up is relative to D4RT.}
    \centering
    \setlength{\tabcolsep}{4pt}
    \resizebox{\textwidth}{!}{%
    \begin{tabular}{l cc cccccccc}
        \toprule
        \multirow{3}{*}{\textbf{Method}}
            & \multicolumn{2}{c}{\textbf{Point Cloud}} & \multicolumn{8}{c}{\textbf{Video Depth}} \\
        \cmidrule(lr){2-3} \cmidrule(lr){4-11}
            & \textbf{Sintel} & \textbf{ScanNet}
            & \multicolumn{2}{c}{\textbf{Sintel}} & \multicolumn{2}{c}{\textbf{ScanNet}}
            & \multicolumn{2}{c}{\textbf{KITTI}} & \multicolumn{2}{c}{\textbf{Bonn}} \\
        \cmidrule(lr){2-2} \cmidrule(lr){3-3} \cmidrule(lr){4-5} \cmidrule(lr){6-7} \cmidrule(lr){8-9} \cmidrule(lr){10-11}
            & L1$\downarrow$ & L1$\downarrow$
            & AbsRel (S)$\downarrow$ & AbsRel (SS)$\downarrow$
            & AbsRel (S)$\downarrow$ & AbsRel (SS)$\downarrow$
            & AbsRel (S)$\downarrow$ & AbsRel (SS)$\downarrow$
            & AbsRel (S)$\downarrow$ & AbsRel (SS)$\downarrow$ \\
        \midrule
        D4RT & 0.997 & 0.250 & 0.392 & 0.307 & 0.069 & 0.065 & 0.097 & 0.091 & 0.071 & 0.059 \\
        OpenD4RT                  & 0.996 & 0.250    & 0.393 & 0.307 & 0.070    & 0.066    & 0.099 & 0.092 & 0.071 & 0.059 \\
        \textbf{\method (Ours)}   & 0.995 & 0.249 & 0.394 & 0.307 & 0.070 & 0.065 & 0.098 & 0.093 & 0.071 & 0.059 \\
        \midrule
        \specialrule{\lightrulewidth}{0pt}{0pt}
        \rowcolor{oursPurple}
        \textbf{Speed-up}\rule[-0.9ex]{0pt}{3ex} & \textbf{6.1$\times$} & \textbf{6.2$\times$}
            & \multicolumn{2}{c}{\textbf{6.1$\times$}} & \multicolumn{2}{c}{\textbf{6.2$\times$}}
            & \multicolumn{2}{c}{\textbf{6.3$\times$}} & \multicolumn{2}{c}{\textbf{6.2$\times$}} \\
        \specialrule{\heavyrulewidth}{0pt}{0pt}
    \end{tabular}}
    \label{tab:dense}
\end{table*}

% \begin{table*}[t]
%     \centering
%     \small
%     \begin{tabular}{l ccc ccc c}
%         \toprule
%         & \multicolumn{3}{c}{Sintel} & \multicolumn{3}{c}{ScanNet} & Re10K \\
%         \cmidrule(lr){2-4} \cmidrule(lr){5-7} \cmidrule(lr){8-8}
%         Method & ATE$\downarrow$ & RPE-T$\downarrow$ & RPE-R$\downarrow$ & ATE$\downarrow$ & RPE-T$\downarrow$ & RPE-R$\downarrow$ & AUC@30$\uparrow$ \\
%         \midrule
%         D4RT & 0.145 & 0.053 & 0.724 & 0.149 & 0.056 & 2.159 & 67.9 \\
%         OpenD4RT & 0.145 & 0.053 & 0.720 & 0.149 & 0.056 & 2.157 & 68.0 \\
%         \textbf{\method (Ours)} & \textbf{0.141} & \textbf{0.039} & \textbf{0.604} & \textbf{0.152} & \textbf{0.068} & \textbf{3.503} & \textbf{68.6} \\
%         \midrule
%         Speed-up & \multicolumn{3}{c}{\textbf{6.0$\times$}} & \multicolumn{3}{c}{\textbf{3.1$\times$}} & \textbf{2.0$\times$} \\
%         \bottomrule
%     \end{tabular}
%     \caption{\textbf{Camera pose estimation} -- We evaluate \method against D4RT on static indoor scenes (ScanNet, Re10K) and dynamic outdoor scenes (Sintel). \method accelerates D4RT on all datasets and improves its accuracy on Sintel. RPE-R is in degrees, and speed-up is relative to D4RT.}
%     \label{tab:pose}
% \end{table*}

\begin{table}[t]
    \caption{\textbf{Camera pose estimation.} \method accelerates D4RT on all datasets and improves its accuracy on Sintel. RPE-R is in degrees, and speed-up is relative to D4RT.}
    \centering
    \setlength{\tabcolsep}{3pt}
    \resizebox{\linewidth}{!}{%
    \begin{tabular}{l ccc ccc c}
        \toprule
        \multirow{2}{*}{\textbf{Method}}
            & \multicolumn{3}{c}{\textbf{Sintel}} & \multicolumn{3}{c}{\textbf{ScanNet}} & \textbf{Re10K} \\
        \cmidrule(lr){2-4} \cmidrule(lr){5-7} \cmidrule(lr){8-8}
            & ATE$\downarrow$ & RPE-T$\downarrow$ & RPE-R$\downarrow$ & ATE$\downarrow$ & RPE-T$\downarrow$ & RPE-R$\downarrow$ & AUC@30$\uparrow$ \\
        \midrule
        D4RT & 0.145 & 0.053 & 0.724 & 0.149 & 0.056 & 2.159 & 67.9 \\
        OpenD4RT                  & 0.145 & 0.053 & 0.720 & 0.149 & 0.056 & 2.157 & 68.0 \\
        \textbf{\method (Ours)}   & 0.141 & 0.039 & 0.604 & 0.152 & 0.068 & 3.503 & 68.6 \\
        \midrule
        \specialrule{\lightrulewidth}{0pt}{0pt}
        \rowcolor{oursPurple}
        \textbf{Speed-up}\rule[-0.9ex]{0pt}{3ex} & \multicolumn{3}{c}{\textbf{6.0$\times$}} & \multicolumn{3}{c}{\textbf{3.1$\times$}} & \textbf{2.1$\times$} \\
        \specialrule{\heavyrulewidth}{0pt}{0pt}
    \end{tabular}}
    \label{tab:pose}
\end{table}

\subsection{Experiment Setup}
\label{sec:setup}

\noindent\textbf{Benchmarks.} We benchmark \method against D4RT on all of its tasks, following the benchmarks and metrics of D4RT~\cite{zhang2025d4rt}. For 4D reconstruction and tracking, we evaluate on the three subsets of TAPVid-3D~\cite{koppula2024tapvid3d}: (i) DriveTrack, (ii) ADT and (iii) PStudio, using the official query points on full-length sequences. We also measure the tracking throughput following D4RT. For dense prediction, we evaluate point cloud estimation on Sintel~\cite{butler2012sintel} and ScanNet~\cite{dai2017scannet}, and video depth estimation on Sintel, ScanNet, KITTI~\cite{geiger2012kitti} and Bonn~\cite{palazzolo2019refusion}. For camera pose estimation, we evaluate on Sintel, ScanNet and RealEstate10K~\cite{zhou2018stereo}.  Details of the datasets and metrics are provided in Appendix~\ref{sec:supp_impl}.

\noindent\textbf{Baselines.} We compare \method with two baselines. 
OpenD4RT is an open-source community reproduction of the full architecture and training recipe of D4RT and reports the strongest result among recent 4D tracking methods.
Since the official implementation and checkpoint of D4RT are not public, we further implement D4RT ourselves on top of the same checkpoint, following the inference protocol of the original paper and adding engineering optimizations for efficient inference. The two baselines achieve nearly identical accuracy, so our optimizations do not affect the accuracy of the community code. Our implementation of D4RT will be fully open source together with \method. In the rest of this paper, D4RT refers to our implementation, and all speed-ups of \method are computed relative to it.

\noindent\textbf{Implementation Details.} \method uses the same hyperparameters across all datasets. All experiments run on NVIDIA H20 GPUs. Further details and hyperparameter sensitivity are provided in Appendix~\ref{sec:supp_hyper} and~\ref{app:sensitivity}.

\subsection{4D Reconstruction and Tracking}
\label{sec:exp_4d}

\noindent\textbf{3D Point Tracking.} Following D4RT, we evaluate 3D point tracking on TAPVid-3D in both camera and world coordinates, with and without ground-truth intrinsics. As shown in Tab.~\ref{tab:tracking}, \method achieves a substantial speedup on all three subsets while maintaining the accuracy of D4RT. On longer videos, \method is also more accurate than D4RT, since encoding fewer segments reduces the error accumulated across them. On DriveTrack, \method remains on par with D4RT, with a slightly lower occlusion accuracy.
The intrinsics of Ground-truth improve both methods, but remain below the results reported in D4RT, which are obtained with the official checkpoint that is not public. Since \method and D4RT share the same OpenD4RT checkpoint, this gap does not affect the comparison between them. 

\noindent\textbf{Tracking Throughput.} Following D4RT, we measure the maximum number of full-video 3D tracks that can be produced from a 48-frame clip while maintaining a target frame rate. As shown in Tab.~\ref{tab:throughput}, \method produces about five times more tracks than D4RT at every frame rate. Since only a single clip is encoded, this gain comes entirely from the decoder, where temporal anchors reduce the queries of each track and shared-token decoding reduces the cost of each query.

Overall, these results demonstrate that \method preserves the tracking ability of D4RT, while producing tracks at a fraction of its cost.

\begin{figure}[t]
    \centering
    \includegraphics[width=\linewidth]{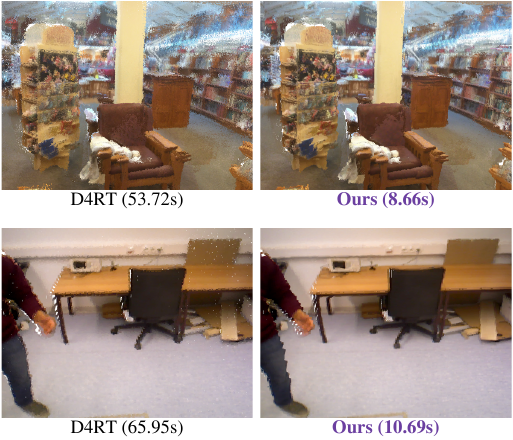}
    \caption{\textbf{Qualitative comparison of dense 3D reconstruction} on a static scene from ScanNet (top) and a dynamic scene from Bonn (bottom). For Bonn, we show the point map of a single frame to avoid ghosting from moving objects. The runtime is measured respectively on the whole sequence of 90 and 110 frames.}
    \label{fig:qual_dense}
\end{figure}

\subsection{3D Reconstruction}
\label{sec:exp_3d}

\noindent\textbf{Point Cloud and Video Depth.} We evaluate point cloud estimation on Sintel and ScanNet, and video depth estimation on Sintel, ScanNet, KITTI and Bonn. As shown in Tab.~\ref{tab:dense}, \method runs about six times faster than D4RT with almost identical accuracy. Notably, this speed-up comes entirely from the decoder, which dominates the cost of dense prediction, so the reduction in queries and attention cost translates almost directly into speed. Qualitative results in Fig.~\ref{fig:qual_dense} further demonstrate this. On both the static and the dynamic scene, \method recovers the same scene geometry and fine details as D4RT, while taking only about one sixth of its runtime. We also note that the community checkpoint is weaker than the official model on Sintel, which affects both methods equally.

\noindent\textbf{Camera Pose Estimation.} We evaluate camera pose on Sintel, ScanNet and RealEstate10K. As shown in Tab.~\ref{tab:pose}, \method consistently accelerates D4RT on all three datasets while maintaining comparable accuracy. On Sintel, it also lowers errors, since interpolation between anchors smooths frame-to-frame jitter. On ScanNet, the relative errors increase slightly, as the handheld camera rotates faster than the anchors can follow. On RealEstate10K, each video has only ten frames, so the speedup is smaller.

These results demonstrate that \method extends to dense prediction and camera pose estimation, and that its speedup is largest where decoding dominates the cost.

\newcommand{\hl}[1]{{\setlength{\fboxsep}{1.5pt}\colorbox{oursPurple}{#1}}}
\begin{table}[t]
    \caption{\textbf{Long-Video Tracking on ADT.} The 39 ADT videos are truncated to their first 64, 128, 192 and 300 frames. AJ, APD$_{3D}$ and OA are in \%. Speed-up is relative to D4RT at the same length.}
    \centering
    \footnotesize
    \setlength{\tabcolsep}{3pt}
    \resizebox{\columnwidth}{!}{%
    \begin{tabular}{c l cccc ccc}
        \toprule
        \multirow{2}{*}{\textbf{Frames}} & \multirow{2}{*}{\textbf{Method}}
            & \multicolumn{4}{c}{\textbf{Camera frame}} & \multicolumn{3}{c}{\textbf{World frame}} \\
        \cmidrule(lr){3-6} \cmidrule(lr){7-9}
        & & AJ$\uparrow$ & APD$_{3D}\uparrow$ & OA$\uparrow$ & Speed-up & APD$_{3D}\uparrow$ & L1$\downarrow$ & Speed-up \\
        \midrule
        \multirow{3}{*}{64}
        & D4RT     & 10.9 & 15.3 & 92.4 & 1.0$\times$ & 68.7 & 0.516 & 1.0$\times$ \\
        & OpenD4RT & 10.8 & 15.2 & 92.4 & 0.2$\times$ & 68.6 & 0.515 & 0.2$\times$ \\
        & \textbf{\method (Ours)} & 15.6 & 21.0 & 90.9 & \hl{\textbf{2.7$\times$}} & 70.7 & 0.543 & \hl{\textbf{3.4$\times$}} \\
        \midrule
        \multirow{3}{*}{128}
        & D4RT     & 5.8 & 9.6 & 85.5 & 1.0$\times$ & 57.3 & 0.794 & 1.0$\times$ \\
        & OpenD4RT & 5.9 & 9.6 & 85.6 & 0.3$\times$ & 57.2 & 0.796 & 0.2$\times$ \\
        & \textbf{\method (Ours)} & 8.7 & 13.3 & 85.8 & \hl{\textbf{3.6$\times$}} & 62.9 & 0.652 & \hl{\textbf{4.1$\times$}} \\
        \midrule
        \multirow{3}{*}{192}
        & D4RT     & 3.9 & 6.9 & 81.5 & 1.0$\times$ & 47.1 & 1.128 & 1.0$\times$ \\
        & OpenD4RT & 3.9 & 7.0 & 81.5 & 0.2$\times$ & 47.0 & 1.131 & 0.2$\times$ \\
        & \textbf{\method (Ours)} & 6.4 & 10.2 & 81.9 & \hl{\textbf{3.8$\times$}} & 52.3 & 0.905 & \hl{\textbf{4.0$\times$}} \\
        \midrule
        \multirow{3}{*}{300}
        & D4RT     & 2.3 & 4.7 & 77.3 & 1.0$\times$ & 34.8 & 1.785 & 1.0$\times$ \\
        & OpenD4RT & 2.3 & 4.7 & 77.3 & 0.2$\times$ & 34.9 & 1.784 & 0.3$\times$ \\
        & \textbf{\method (Ours)} & 3.0 & 5.5 & 78.0 & \hl{\textbf{5.3$\times$}} & 39.3 & 1.564 & \hl{\textbf{5.8$\times$}} \\
        \bottomrule
    \end{tabular}}
    \label{tab:long}
\end{table}

\subsection{Long-Video Tracking}
\label{sec:long_video}

\begin{table*}[t]
    \centering
    \setlength{\tabcolsep}{3.2pt}
    \resizebox{\textwidth}{!}{%
    \begin{tabular}{l ccc cc cc ccc cccc}
        \toprule
        \multirow{3}{*}{\textbf{Method}}
            & \multicolumn{3}{c}{\textbf{3D track (camera)}} & \multicolumn{2}{c}{\textbf{3D track (world)}} & \multicolumn{2}{c}{\textbf{Point cloud}} & \multicolumn{3}{c}{\textbf{Video depth}} & \multicolumn{4}{c}{\textbf{Camera pose}} \\
        \cmidrule(lr){2-4} \cmidrule(lr){5-6} \cmidrule(lr){7-8} \cmidrule(lr){9-11} \cmidrule(lr){12-15}
            & DriveTrack & ADT & PStudio & DriveTrack & ADT & Sintel & ScanNet & Sintel & ScanNet & Bonn & Sintel & \multicolumn{2}{c}{ScanNet} & Re10K \\
        \cmidrule(lr){13-14}
            & AJ$\uparrow$ & AJ$\uparrow$ & AJ$\uparrow$ & APD$\uparrow$ & APD$\uparrow$ & L1$\downarrow$ & L1$\downarrow$ & AbsRel$\downarrow$ & AbsRel$\downarrow$ & AbsRel$\downarrow$ & ATE$\downarrow$ & ATE$\downarrow$ & RPE-R$\downarrow$ & AUC@30$\uparrow$ \\
        \midrule
        \multicolumn{15}{l}{\textit{Accuracy}} \\
        D4RT              & 1.24 & 2.26 & 6.47 & 3.10 & 34.79 & 0.997 & 0.250 & 0.392 & 0.070 & 0.071 & 0.145 & 0.149 & 2.159 & 67.9 \\
        w/o ASI           & 1.21 & 2.25 & 5.41 & 3.07 & 34.74 & 0.994 & 0.249 & 0.392 & 0.070 & 0.071 & 0.144 & 0.149 & 2.148 & 68.4 \\
        w/o STD           & 1.30 & 3.03 & 7.47 & 3.18 & 39.47 & 0.997 & 0.250 & 0.393 & 0.070 & 0.071 & 0.142 & 0.153 & 3.503 & 68.5 \\
        \textbf{\method}  & 1.25 & 3.01 & 6.61 & 3.16 & 39.30 & 0.995 & 0.249 & 0.394 & 0.070 & 0.071 & 0.141 & 0.152 & 3.503 & 68.6 \\
        \midrule
        \multicolumn{15}{l}{\textit{Speed-up over D4RT}} \\
        w/o ASI           & 1.2$\times$ & 1.4$\times$ & 1.2$\times$ & 1.3$\times$ & 1.4$\times$ & 1.8$\times$ & 1.7$\times$ & 1.9$\times$ & 1.7$\times$ & 1.7$\times$ & 1.5$\times$ & \multicolumn{2}{c}{1.5$\times$} & 1.3$\times$ \\
        w/o STD           & 2.6$\times$ & 5.1$\times$ & 4.4$\times$ & 3.2$\times$ & 5.5$\times$ & 4.5$\times$ & 4.6$\times$ & 4.5$\times$ & 4.6$\times$ & 4.6$\times$ & 6.1$\times$ & \multicolumn{2}{c}{3.0$\times$} & 2.0$\times$ \\
        \rowcolor{oursPurple}
        \textbf{\method}\rule[-0.9ex]{0pt}{3ex}  & \textbf{2.8$\times$} & \textbf{5.3$\times$} & \textbf{4.2$\times$} & \textbf{3.1$\times$} & \textbf{5.8$\times$} & \textbf{6.1$\times$} & \textbf{6.2$\times$} & \textbf{6.1$\times$} & \textbf{6.2$\times$} & \textbf{6.2$\times$} & \textbf{6.0$\times$} & \multicolumn{2}{c}{\textbf{3.1$\times$}} & \textbf{2.1$\times$} \\
        \specialrule{\heavyrulewidth}{0pt}{0pt}
    \end{tabular}}
    \caption{\textbf{Ablation of components on all benchmarks.} ASI: anchor-based sparse inference. STD: shared-token decoding. Each variant removes one component of \method.}
    \label{tab:ablation}
\end{table*}

\begin{figure*}[t]
    \centering
    \includegraphics[width=0.95\linewidth]{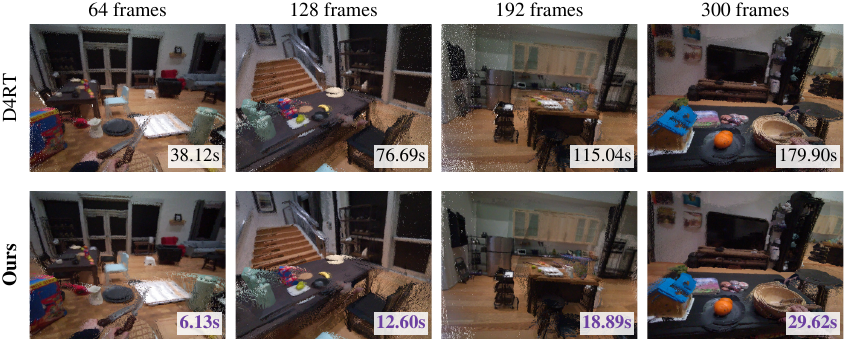}
    \caption{\textbf{Qualitative comparison of dense 3D reconstruction on long videos.} Each column shows a different ADT sequence truncated to 64, 128 192 and 300 frames. The runtime of each method is shown at the bottom right. \method produces point clouds nearly identical to those of D4RT at about 6$\times$ lower runtime.}
    \label{fig:qual_long}
\end{figure*}

To examine how both methods scale with the length of the video, we truncate the ADT videos to their first 64, 128, 192 and 300 frames, so that each length requires a different number of segments.

As shown in Tab.~\ref{tab:long}, the accuracy of D4RT drops quickly as videos get longer, since more segments are chained and every additional alignment introduces error. In contrast, \method encodes fewer segments at every length, and achieves higher AJ and APD$_{3D}$ than D4RT in all settings. The speedup also grows with the length of the video, from about three times at 64 frames to more than five times at 300 frames. At 64 frames, where the whole video fits into one clip after frame selection, \method shows a slightly lower occlusion accuracy, since the skipped frames are recovered by interpolation rather than decoded. These results demonstrate that \method not only accelerates D4RT, but also mitigates error accumulation on long videos. 
Qualitative results in Fig.~\ref{fig:qual_long} further demonstrate that \method achieves reconstruction quality comparable to D4RT at every length, while running about six times faster. This speed-up stays constant as the video grows, since dense outputs do not subsample the encoder input, and the gain comes entirely from decoding fewer queries at a lower cost per query.

\subsection{Ablation Study}
\label{sec:ablation}

To isolate the contribution of each component, we construct two variants of \method: (i) \method (w/o ASI), which decodes every query and encodes every frame, while keeping shared-token decoding unchanged; and (ii) \method (w/o STD), which lets all queries attend to all tokens in every decoder block, while keeping anchor-based sparse inference unchanged. We evaluate both variants and \method on all benchmarks of D4RT.

The results in Tab.~\ref{tab:ablation} show that both components contribute to acceleration. Removing anchor-based sparse inference or shared-token decoding alone reduces the speedup on most tasks, and anchor-based sparse inference also accounts for the accuracy gain on long videos, since fewer segments are chained. The gain of shared-token decoding is largest on dense prediction, where many queries share one token set, and small on 3D tracking, where fewer queries per segment amortize the cost of token selection. \method achieves the highest speedup in most tasks, supporting our claim that much of the computation in D4RT is redundant. See additional experiments and analyzes in Appendix~\ref{sec:supp_results}.

\section{Conclusion}
\label{sec:conclusion}
\vspace{-0.4mm}
In this work, we proposed \method, a training-free framework that accelerates D4RT. By analyzing the inference, we found that it encodes every frame and decodes every query densely, even though most of this computation is redundant. In particular, most outputs change smoothly, and the later decoder blocks attend to a small set of tokens shared by all queries. To address this, \method encodes and decodes only at sparse anchors and recovers the rest by interpolation, and lets the remaining queries attend only to a shared set of tokens. Experiments on the benchmarks of D4RT show that \method achieves up to 6.2 times speedup while maintaining competitive accuracy, and further mitigates error accumulation on long videos. These results suggest that much of the cost of query-based 4D reconstruction can be removed at test time, bringing such models closer to real-time use.

{
    \small
    \bibliographystyle{ieeenat_fullname}
    \bibliography{main}
}

% Appendix is kept in the arXiv version, after the references
\clearpage
\appendix
\renewcommand\maketitlesupplementary{%
  \onecolumn
  \begin{center}
    {\Large\textbf{\thetitle}}\\[0.5em]
    {\Large Appendix}
  \end{center}
  \vspace{1.0em}}

\maketitlesupplementary
\appendix
\setcounter{page}{1}
\section{Related Work}
\label{sec:related}

\subsection{Feed-Forward 3D and 4D Reconstruction}

Classical pipelines recover geometry from images with Structure-from-Motion and Multi-View Stereo~\cite{schoenberger2016sfm,schoenberger2016mvs}. They are accurate on static scenes, but they rely on per-scene optimization, which is slow and fragile on dynamic videos. DUSt3R~\cite{wang2024dust3r} changed this by regressing point maps of an image pair with a single network, and MASt3R~\cite{leroy2024mast3r} further added dense matching. Later models extend this design to many views in one pass. Fast3R~\cite{yang2025fast3r} processes over a thousand images at once, VGGT~\cite{wang2025vggt} alternates frame-wise and global attention to predict cameras, depth and point maps jointly, and $\pi^3$~\cite{wang2025pi3} removes the dependence on a fixed reference view.

For dynamic scenes, MonST3R~\cite{zhang2025monst3r} fine-tunes DUSt3R on dynamic data, CUT3R~\cite{wang2025cut3r} keeps a persistent state to process a video online, and Easi3R~\cite{chen2025easi3r} reuses the attention of DUSt3R to separate motion without training. Other methods combine several models for depth, motion and camera~\cite{li2025megasam,lu2025align3r,wang2025c4d}. However, these models predict a dense output for every input frame, so they do not directly give the correspondence of a dynamic point over time. D4RT~\cite{zhang2025d4rt} instead follows the query-based design of scene representation transformers~\cite{sajjadi2022srt,jaegle2022perceiverio}. It encodes a video once and obtains every 4D output by querying the scene representation, which unifies depth, point tracks and camera parameters in one model. FastD4RT is built on this query-based design. 

\subsection{3D Point Tracking}

Point tracking estimates the motion of a query point through a video. Particle Video~\cite{sand2008particle} first tracked points over long ranges, and learning-based trackers~\cite{harley2022pips,doersch2022tapvid,doersch2023tapir,karaev2024cotracker} have made 2D tracking robust to occlusions. Dense trackers such as DOT~\cite{lemoing2024dot} further track every pixel of a frame. A 2D track can be lifted to 3D with a depth map and camera intrinsics, as in the baselines of TAPVid-3D~\cite{koppula2024tapvid3d}, but the result depends on the accuracy of each component.

Recent methods predict 3D tracks directly. SpatialTracker~\cite{xiao2024spatialtracker,xiao2025spatialtrackerv2} tracks points in 3D space and later models camera motion explicitly, DELTA~\cite{ngo2025delta} tracks every pixel in 3D efficiently, and L4P~\cite{badki2026l4p} shares a video backbone across low-level 4D tasks. St4RTrack~\cite{feng2025st4rtrack} extend the pairwise point maps of DUSt3R to reconstruction and tracking at the same time. D4RT obtains a 3D track by querying the same point at every target time, and proposes an algorithm that exploits spatio-temporal redundancy to track all pixels with fewer queries~\cite{zhang2025d4rt}. FastD4RT keeps this tracking interface unchanged. It decodes each track only at temporal anchors and recovers the remaining frames by interpolation.

\subsection{Efficient Inference for 3D Vision Models}

As feed-forward 3D models~\cite{yang2025fast3r,wang2025pi3,wang2025cut3r} scale to long sequences, their inference cost has become a key concern. Most existing work accelerates the global attention of VGGT-style models~\cite{wang2025vggt}. FastVGGT~\cite{shen2025fastvggt} observes that the global attention of VGGT collapses to similar patterns and merges redundant tokens. HTTM~\cite{wang2026httm} merges tokens separately in each attention head and exploits temporal correspondence, and block-sparse attention~\cite{wang2025fastervggt,dao2022flashattention} computes only the blocks of the global attention map where the attention mass concentrates. StreamVGGT~\cite{zhuo2025streamvggt} turns the global attention into causal attention with a key--value cache, so video can be processed frame by frame. For videos longer than the model can take at once, VGGT-Long~\cite{deng2026vggt} splits the input into overlapping chunks and aligns them, which is also the protocol D4RT uses for long videos.

These methods reduce the cost of building the scene representation in models that predict a dense output for every frame. In contrast, the cost of D4RT is split between encoding many segments and decoding a large number of independent queries, and the decoder alone accounts for almost all of the runtime in dense prediction. FastD4RT therefore targets both the number of encoded segments and decoded queries and the cost of each query. Since encoder acceleration methods~\cite{rao2021dynamicvit,liang2022evit} change only the attention inside the encoder, they are orthogonal to FastD4RT and could be combined with it.

\subsection{Token Reduction and Sparse Attention}

Reducing the tokens that attention operates on is a common way to accelerate transformers~\cite{child2019sparse,beltagy2020longformer,zaheer2020bigbird}. In vision transformers, DynamicViT~\cite{rao2021dynamicvit} and EViT~\cite{liang2022evit} prune or fuse less informative tokens~\cite{yin2022avit,meng2022adavit,xu2022evovit}, and ToMe~\cite{bolya2023tome,bolya2023tomesd} merges similar tokens without training. In large language models, the key--value cache is compressed by keeping only the tokens that receive most attention~\cite{liu2023scissorhands,ge2024fastgen,tang2024quest}. StreamingLLM~\cite{xiao2024streamingllm} keeps a few initial tokens as attention sinks, and SnapKV~\cite{li2024snapkv} selects important tokens for each head from the attention of a small observation window. Shared-token decoding follows the same principle of keeping the tokens that receive most attention. Unlike these methods, it selects the token set once per segment from a small set of calibration queries, and shares this set across all queries and tasks of the decoder, since the later decoder blocks of D4RT attend to almost the same tokens for every query.

Computing expensive outputs only at sparse locations is also a long-standing idea in video analysis~\cite{shelhamer2016clockwork,jain2019accel}. Deep Feature Flow~\cite{zhu2017dff}, for example, runs the full network only on sparse key frames and propagates their features to the other frames~\cite{zhu2018towards}. D4RT already estimates camera parameters from a coarse grid of queries~\cite{zhang2025d4rt}. FastD4RT extends this idea to all outputs of D4RT and to the frames given to the encoder, and bounds the error introduced by interpolation.

%% ============================================================
\section{Theoretical Analysis}
\label{sec:supp_theory}
\subsection{Proof of Theorem~\ref{thm:interp}}
\label{sec:supp_proof_interp}

\noindent\textbf{Interpolation error.} We first bound the error of linear interpolation between two anchors $a<b$, with spacing $\rho=b-a$. Let $\mathbf{y}$ be twice differentiable on $[a,b]$ with $\|\mathbf{y}''\|\le\kappa$, and let $\hat{\mathbf{y}}$ be its linear interpolation in Eq.~\eqref{eq:interp}. For any unit vector $\mathbf{w}$, let $\phi(t)=\mathbf{w}^\top(\mathbf{y}(t)-\hat{\mathbf{y}}(t))$ and $\psi(t)=\frac{\kappa}{2}(t-a)(b-t)$. Since $\hat{\mathbf{y}}$ is linear, $|\phi''|=|\mathbf{w}^\top\mathbf{y}''|\le\kappa$, while $\psi''=-\kappa$. Hence $\psi-\phi$ and $\psi+\phi$ are concave, and both vanish at $a$ and $b$, so they are non-negative on $[a,b]$. This gives $|\phi(t)|\le\psi(t)\le\frac{\rho^2}{8}\kappa$. Taking $\mathbf{w}$ along $\mathbf{y}(t)-\hat{\mathbf{y}}(t)$,
\begin{equation}
    \|\mathbf{y}(t)-\hat{\mathbf{y}}(t)\| \le \frac{\rho^2}{8}\max\|\mathbf{y}''\|.
    \label{eq:supp_linear}
\end{equation}
Dense outputs are interpolated along $u$ and then along $v$. Let $\bar{\mathbf{y}}$ be $\mathbf{y}$ interpolated along $u$ only, so that $\hat{\mathbf{y}}$ is $\bar{\mathbf{y}}$ interpolated along $v$, and
\begin{equation}
    \mathbf{y}-\hat{\mathbf{y}} = (\mathbf{y}-\bar{\mathbf{y}}) + (\bar{\mathbf{y}}-\hat{\mathbf{y}}).
    \label{eq:supp_bilinear}
\end{equation}
The first term is bounded by Eq.~\eqref{eq:supp_linear} along $u$. Since $\bar{\mathbf{y}}$ is a convex combination of $\mathbf{y}$ at two neighboring grid columns, its second derivative along $v$ is bounded by that of $\mathbf{y}$, so the second term is bounded by Eq.~\eqref{eq:supp_linear} along $v$. The error of bilinear interpolation is therefore the sum of the errors along $u$ and $v$.

\noindent\textbf{3D tracks.} A world-frame track is interpolated along time with anchor spacing $\rho=r$. Its second derivative is the acceleration of the point, which is at most $\alpha$ under Assumption~\ref{assu:smooth}. By Eq.~\eqref{eq:supp_linear},
\begin{equation}
    E_{\text{int}} \le \frac{r^2}{8}\max\|\mathbf{y}''\| \le \frac{r^2}{8}\alpha.
    \label{eq:supp_track}
\end{equation}
Rotations of camera poses are interpolated along the geodesic, which is linear in the tangent space around the first anchor, so the same argument applies there. When the camera rotates with a constant angular velocity, the rotation follows the geodesic and is interpolated exactly.

\noindent\textbf{Dense outputs.} Let $\mathbf{p}$ be the ray of pixel $(u,v)$ with unit depth, which is affine in $(u,v)$ for a pinhole camera, so that the 3D point is $\mathbf{X}=z\,\mathbf{p}$ with depth $z$. Under Assumption~\ref{assu:smooth}, the surface is a plane $\mathbf{n}^\top\mathbf{X}=1$, so the inverse depth $\frac{1}{z} = \mathbf{n}^\top\mathbf{p}$
is also affine in $(u,v)$. Differentiating $z=1/(\mathbf{n}^\top\mathbf{p})$ twice along $u$ gives
\begin{equation}
    \partial_u^2 z = \frac{2(\partial_u z)^2}{z}.
    \label{eq:supp_depth}
\end{equation}
Neighboring grid points are $\rho=W/G$ apart along $u$, so the relative depth difference between them is $\Delta=\rho|\partial_u z|/z$, and the depth difference is $\ell=\rho|\partial_u z|$. Eq.~\eqref{eq:supp_depth} then reads
\begin{equation}
    |\partial_u^2 z| = \frac{2\Delta\ell}{\rho^2}.
    \label{eq:supp_curv}
\end{equation}
For point clouds, since $\mathbf{p}$ is affine in $u$, its second derivative vanishes, and
\begin{equation}
    \partial_u^2\mathbf{X} = (\partial_u^2 z)\,\mathbf{p} + 2(\partial_u z)\,\partial_u\mathbf{p} = \frac{2\,\partial_u z}{z}\big((\partial_u z)\,\mathbf{p}+z\,\partial_u\mathbf{p}\big) = \frac{2\,\partial_u z}{z}\,\partial_u\mathbf{X}.
    \label{eq:supp_pc}
\end{equation}
Here $\ell=\rho\|\partial_u\mathbf{X}\|$ is the 3D distance between neighboring grid points, so $\|\partial_u^2\mathbf{X}\|=2\Delta\ell/\rho^2$, the same as Eq.~\eqref{eq:supp_curv}. For both outputs, Eq.~\eqref{eq:supp_linear} bounds the error along each axis by $\frac{1}{4}\Delta\ell$, and Eq.~\eqref{eq:supp_bilinear} adds the two axes:
\begin{equation}
    E_{\text{int}} \le 2\cdot\frac{\rho^2}{8}\cdot\frac{2\Delta\ell}{\rho^2} = \frac{1}{2}\Delta\ell.
    \label{eq:supp_dense}
\end{equation}
A rigid transform to the world frame preserves this bound. Finally, both $\Delta$ and $\ell$ are proportional to the grid spacing $H/G$, so $E_{\text{int}}=\mathcal{O}\big((H/G)^2\big)$ for dense outputs, as stated in Eq.~\eqref{eq:takeaways}.

\subsection{Proof of Theorem~\ref{thm:shared}}
\label{sec:supp_proof_shared}

Let $\mathbf{a}$ be the attention of a query over the $N$ tokens in a head, as in Eq.~\eqref{eq:shared}, and let $\epsilon$ be its total weight outside the shared tokens. Let $\bar{\mathbf{v}}_{\text{in}}$ and $\bar{\mathbf{v}}_{\text{out}}$ be the attention-weighted means of the values inside and outside the shared tokens. The full attention output is
\begin{equation}
    \mathbf{o} = (1-\epsilon)\,\bar{\mathbf{v}}_{\text{in}} + \epsilon\,\bar{\mathbf{v}}_{\text{out}}.
\end{equation}
Restricting the attention to the shared tokens renormalizes the weights inside them, so $\tilde{\mathbf{o}}=\bar{\mathbf{v}}_{\text{in}}$. Therefore,
\begin{equation}
    \tilde{\mathbf{o}}-\mathbf{o} = \epsilon\,(\bar{\mathbf{v}}_{\text{in}}-\bar{\mathbf{v}}_{\text{out}}).
\end{equation}
Both means are convex combinations of the values, so their norms are at most $V$, which gives $E_{\text{sh}}\le 2\epsilon V$. The error is zero when no attention falls outside the shared tokens, and small when the tokens outside carry values similar to those inside.

Finally, linear and bilinear interpolation form convex combinations of the anchors, so the error of the interpolated outputs never exceeds the largest error at the anchors. Hence interpolation does not amplify the shared-token error.

\section{Framework}
\label{app:framework}
FastD4RT takes a video $\mathcal{V}$ of $T$ frames and the frozen encoder $\mathcal{E}$ and decoder $\mathcal{D}$ of D4RT, and runs as in Algorithm~\ref{alg:fastd4rt}. If the video is longer than the training clip length $T_c$ and the output is not dense, FastD4RT keeps every $s$-th frame. The video is then split into $K$ overlapping segments, and each segment $\mathcal{S}_k$ is encoded once into $\mathbf{F}_k$, from which the keys and values of all decoder blocks are computed and cached.

For each segment, FastD4RT decodes the calibration queries $\mathcal{C}$ with full attention, and keeps the $m$ tokens with the largest mean attention as the shared set $\mathcal{T}_l^h$ of each head $h$ in each block $l>l_0$ (Eq.~\eqref{eq:shared}). It then decodes the anchor queries in parallel, at every $r$-th frame for tracks and camera poses and on a $G\times G$ grid for dense outputs. Each query attends to all tokens in the first $l_0$ blocks and only to the shared tokens in the rest, and the outputs between anchors are recovered by interpolation (Eq.~\eqref{eq:interp}).

Each segment after the first is registered to its predecessor by a Sim(3) transform, estimated with Umeyama's method~\cite{umeyama1991least} on the most confident points in their overlap, and tracks are handed over through their decoded 2D positions in the overlap. Finally, the transforms are chained to the first segment to give the full-video outputs $\mathcal{Y}$.

Overall, FastD4RT keeps D4RT itself unchanged, and only changes which frames are encoded, which queries are decoded, and which tokens the later blocks attend to.

\begin{algorithm}[t]
\caption{Overview of \method}
\label{alg:fastd4rt}
\begin{algorithmic}[1]
\Statex \textbf{Input:} Video $\mathcal{V}=\{I_t\}_{t=1}^{T}$; encoder $\mathcal{E}$ and decoder $\mathcal{D}$ with $L$ blocks; clip length $T_c$; input stride $s$; anchor spacing $r$; dense grid $G$; full-attention blocks $l_0$; shared tokens $m$; calibration queries $\mathcal{C}$.
\Statex \textbf{Output:} Full-video outputs $\mathcal{Y}$.
\If{$T > T_c$ \textbf{and} the output is not dense}
    \State $\mathcal{V} \gets \textsc{Subsample}(\mathcal{V}; s)$;
\EndIf
\State Split $\mathcal{V}$ into overlapping segments $\{\mathcal{S}_k\}_{k=1}^{K}$;
\For{$k = 1$ \textbf{to} $K$}
    \State Encode segment: $\mathbf{F}_k \gets \mathcal{E}(\mathcal{S}_k)$;
    \State Cache keys and values: $\mathbf{K}_l \gets \mathrm{LN}_l(\mathbf{F}_k)\mathbf{W}_K^l$, $\mathbf{V}_l \gets \mathrm{LN}_l(\mathbf{F}_k)\mathbf{W}_V^l$ for $l=1,\dots,L$;
    \State Decode calibration queries: $\{\mathbf{a}_l^h(\mathbf{q})\}_{\mathbf{q}\in\mathcal{C}} \gets \textsc{FullDecode}(\mathcal{C}; \mathbf{F}_k)$;
    \For{$l = l_0+1$ \textbf{to} $L$}
        \State Select shared tokens: $\mathcal{T}_l^h \gets \textsc{TopM}\big(\frac{1}{|\mathcal{C}|}\sum_{\mathbf{q}\in\mathcal{C}}\mathbf{a}_l^h(\mathbf{q}); m\big)$;
    \EndFor
    \State Build anchor queries: $\mathcal{Q}_k^{a} \gets \textsc{BuildAnchors}(\mathcal{S}_k; r, G)$;
    \For{each query $\mathbf{q} \in \mathcal{Q}_k^{a}$ \textbf{in parallel}}
        \For{$l = 1$ \textbf{to} $L$}
            \If{$l \le l_0$}
                \State $\mathbf{z}_{l+1} \gets \mathrm{CrossAttn}(\mathbf{z}_l, \mathbf{K}_l, \mathbf{V}_l)$;
            \Else
                \State $\mathbf{z}_{l+1} \gets \mathrm{CrossAttn}(\mathbf{z}_l, \mathbf{K}_l[\mathcal{T}_l^h], \mathbf{V}_l[\mathcal{T}_l^h])$;
            \EndIf
        \EndFor
    \EndFor
    \State Recover outputs: $\mathcal{Y}_k \gets \textsc{Interpolate}(\mathcal{Q}_k^{a})$;
    \If{$k > 1$}
        \State Register segments: $\mathbf{T}_k \gets \textsc{Sim3Align}(\mathcal{Y}_{k-1}, \mathcal{Y}_k)$;
    \EndIf
    \If{$k < K$}
        \State Hand over tracks: $\textsc{HandOver}(\mathcal{Y}_k, \mathcal{S}_{k+1})$ at decoded anchors;
    \EndIf
\EndFor
\State Chain segments: $\mathcal{Y} \gets \textsc{Chain}(\{\mathcal{Y}_k\}, \{\mathbf{T}_k\})$;
\State \Return $\mathcal{Y}$
\end{algorithmic}
\end{algorithm}

%% ============================================================
\section{Additional Experiment Setup}
\label{sec:supp_setup}

\subsection{Datasets and Metrics}
\label{sec:supp_impl}

\noindent\textbf{Datasets.} Following D4RT~\cite{zhang2025d4rt}, we evaluate \method on 3D point tracking, point cloud estimation, video depth estimation and camera pose estimation with the following datasets:
\begin{itemize}
    \item \textbf{TAPVid-3D}~\cite{koppula2024tapvid3d}: It contains three subsets. DriveTrack (50 videos) contains outdoor driving scenes from the Waymo Open Dataset, ADT (39 videos) contains egocentric indoor videos from the Aria Digital Twin dataset, and PStudio (50 videos) contains videos of people moving in the Panoptic Studio.
    \item \textbf{Sintel}~\cite{butler2012sintel}: It contains synthetic sequences with fast motion and large camera movement.
    \item \textbf{Bonn}~\cite{palazzolo2019refusion}: It contains real-world dynamic indoor scenes.
    \item \textbf{ScanNet}~\cite{dai2017scannet}: It contains real-world static indoor scenes.
    \item \textbf{KITTI}~\cite{geiger2012kitti}: It contains real-world outdoor driving scenes with sparse depth from LiDAR.
    \item \textbf{RealEstate10K}~\cite{zhou2018stereo}: It contains real-estate videos with SLAM camera poses.
\end{itemize}

\noindent\textbf{Metrics.} We report the following metrics:
\begin{itemize}
    \item \textbf{3D point tracking:} AJ, APD$_{3D}$ and OA in camera coordinates with the official TAPVid-3D code, and APD$_{3D}$ and L1 in world coordinates after a global scale alignment.
    \item \textbf{Video depth:} AbsRel under per-video scale (S) and scale-and-shift (SS) alignment.
    \item \textbf{Point cloud:} L1 distance after centering and normalizing both the predicted and ground-truth clouds.
    \item \textbf{Camera pose:} ATE, RPE-T and RPE-R after Sim(3) alignment, and AUC@30 on RealEstate10K.
    \item \textbf{Tracking throughput:} The largest number of full-video tracks for which encoding a 48-frame clip and decoding all tracks finish within $48/f$ seconds at a target frame rate $f$.
\end{itemize}

\subsection{Hyperparameters}
\label{sec:supp_hyper}

\method uses the same anchor-based sparse inference and shared-token decoding on all tasks. Tracks are decoded at every third frame ($r=3$). For videos longer than one clip, the encoder takes every second frame ($s=2$), and the video is split into segments with a stride of 40 encoded frames. Dense outputs are decoded on a $128\times128$ grid. Camera poses are solved at every third frame on a $16\times16$ grid, with each point weighted by the product of its predicted confidences in the two frames, and videos of at most 72 frames are sampled to one clip.
Shared-token decoding uses full attention in the first $l_0 = 3$ decoder blocks and keeps $m = 512$ tokens per head in the other blocks, selected with $|\mathcal{C}| = 768$ calibration queries per segment.

%% ============================================================
\section{Additional Results}
\label{sec:supp_results}

\subsection{Additional Ablation Study}
\label{app:pertoken}
We compare shared-token decoding with a per-query variant that selects the top-$m$ tokens of each query from its own attention in blocks $l > l_0$. As shown in Tab.~\ref{tab:pertoken}, the two variants reach almost the same accuracy, with per-query selection slightly higher on 3D tracking. However, per-query selection requires the full attention scores of every query and a separate gather of keys and values for each query, so it is only 1.2--3.4$\times$ faster than D4RT, compared with 2.1--6.2$\times$ for shared tokens. Selecting one token set per segment thus keeps the accuracy of per-query selection, while the restricted attention runs as a dense operation over a small common key--value set.

\begin{table*}[t]
    \centering
    \setlength{\tabcolsep}{3.2pt}
    \resizebox{\textwidth}{!}{%
    \begin{tabular}{l ccc cc cc ccc cccc}
        \toprule
        \multirow{3}{*}{\textbf{Method}}
            & \multicolumn{3}{c}{\textbf{3D track (camera)}} & \multicolumn{2}{c}{\textbf{3D track (world)}} & \multicolumn{2}{c}{\textbf{Point cloud}} & \multicolumn{3}{c}{\textbf{Video depth}} & \multicolumn{4}{c}{\textbf{Camera pose}} \\
        \cmidrule(lr){2-4} \cmidrule(lr){5-6} \cmidrule(lr){7-8} \cmidrule(lr){9-11} \cmidrule(lr){12-15}
            & DriveTrack & ADT & PStudio & DriveTrack & ADT & Sintel & ScanNet & Sintel & ScanNet & Bonn & Sintel & \multicolumn{2}{c}{ScanNet} & Re10K \\
        \cmidrule(lr){13-14}
            & AJ$\uparrow$ & AJ$\uparrow$ & AJ$\uparrow$ & APD$\uparrow$ & APD$\uparrow$ & L1$\downarrow$ & L1$\downarrow$ & AbsRel$\downarrow$ & AbsRel$\downarrow$ & AbsRel$\downarrow$ & ATE$\downarrow$ & ATE$\downarrow$ & RPE-R$\downarrow$ & AUC@30$\uparrow$ \\
        \midrule
        \multicolumn{15}{l}{\textit{Accuracy}} \\
        D4RT              & 1.24 & 2.26 & 6.47 & 3.10 & 34.79 & 0.997 & 0.250 & 0.392 & 0.070 & 0.071 & 0.145 & 0.149 & 2.159 & 67.9 \\
        Per-query top-$m$ & 1.30 & 3.05 & 7.48 & 3.18 & 39.41 & 0.996 & 0.250 & 0.393 & 0.070 & 0.071 & 0.142 & 0.152 & 3.503 & 68.5 \\
        \textbf{\method}  & 1.25 & 3.01 & 6.61 & 3.16 & 39.30 & 0.995 & 0.249 & 0.394 & 0.070 & 0.071 & 0.141 & 0.152 & 3.503 & 68.6 \\
        \midrule
        \multicolumn{15}{l}{\textit{Speed-up over D4RT}} \\
        Per-query top-$m$ & 2.0$\times$ & 2.6$\times$ & 3.1$\times$ & 2.0$\times$ & 2.5$\times$ & 1.3$\times$ & 1.2$\times$ & 1.3$\times$ & 1.2$\times$ & 1.2$\times$ & 3.4$\times$ & \multicolumn{2}{c}{1.4$\times$} & 1.7$\times$ \\
        \textbf{\method}  & \textbf{2.8$\times$} & \textbf{5.3$\times$} & \textbf{4.2$\times$} & \textbf{3.1$\times$} & \textbf{5.8$\times$} & \textbf{6.1$\times$} & \textbf{6.2$\times$} & \textbf{6.1$\times$} & \textbf{6.2$\times$} & \textbf{6.2$\times$} & \textbf{6.0$\times$} & \multicolumn{2}{c}{\textbf{3.1$\times$}} & \textbf{2.1$\times$} \\
        \bottomrule
    \end{tabular}}
    \caption{\textbf{Shared vs.\ per-query token selection.} Per-query top-$m$ selects the tokens for each query separately from its own attention, instead of sharing one set per segment.}
    \label{tab:pertoken}
\end{table*}

\subsection{Hyperparameter Sensitivity}
\label{app:sensitivity}
\begin{figure*}[t]
    \centering
    \includegraphics[width=\textwidth]{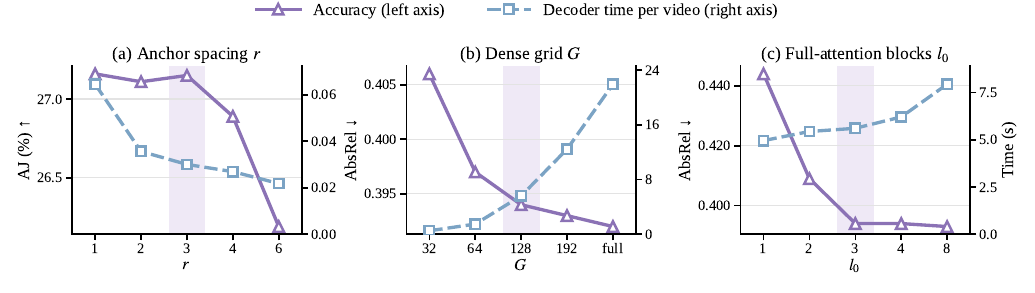}
    \caption{\textbf{Sensitivity to hyperparameters.} Each panel varies one hyperparameter with the others fixed at their defaults (shaded). (a) 3D tracking on ADT. (b, c) Video depth on Sintel. The last point in (b) decodes every pixel, and the last point in (c) disables shared-token decoding. Time is the decoder time per video.}
    \label{fig:sensitivity}
\end{figure*}

We study the sensitivity of \method{} to the anchor spacing $r$, the dense grid size $G$, and the number of full-attention blocks $l_0$, varying one at a time with the others fixed at their defaults. We evaluate $r$ on 3D tracking on the first 48 frames of ADT, and $G$ and $l_0$ on video depth on Sintel. As shown in Fig.~\ref{fig:sensitivity}, accuracy saturates at the default values, while further increases mainly add decoding time. The saturation point of $l_0$ also matches Fig.~\ref{fig:shared_ratio}, where the shared tokens capture almost all of the attention from the fourth block on.

\end{document}